\documentclass[11pt,a4paper]{article}
\usepackage[hyperref]{eacl2021}
\usepackage{times}
\usepackage{latexsym}
\usepackage{graphicx}
\usepackage{amsmath}
\usepackage{multirow}
\usepackage{caption}
\usepackage{subcaption}

\usepackage{microtype}

\aclfinalcopy 
\def\aclpaperid{909} 

\DeclareMathOperator*{\argmax}{arg\,max} 

\title{Event-Driven News Stream Clustering using Entity-Aware Contextual Embeddings}

\author{Kailash Karthik Saravanakumar\thanks{\ \ Work done during internship at Amazon}\ $^{2}$ ~~~~ Miguel Ballesteros$^{1}$ \\ 
  \textbf{Muthu Kumar Chandrasekaran$^{1}$~~~~ Kathleen McKeown$^{1, 2}$} \\
  $^{1}$Amazon AI, USA \\
  $^{2}$Department of Computer Science, Columbia University, NY, USA \\
  \texttt{\{ballemig, cmuthuk, mckeownk\}@amazon.com} \\
  \texttt{\{kailashkarthik.s\}@columbia.edu} \\}

\date{}

\begin{document}
\maketitle
\begin{abstract}
We propose a method for online news stream clustering that is a variant of the non-parametric streaming K-means algorithm. Our model uses a combination of sparse and dense document representations, aggregates document-cluster similarity along these multiple representations and makes the clustering decision using a neural classifier. The weighted document-cluster similarity model is learned using a novel adaptation of the triplet loss into a linear classification objective. We show that the use of a suitable fine-tuning objective and external knowledge in pre-trained transformer models yields significant improvements in the effectiveness of contextual embeddings for clustering. Our model achieves a new state-of-the-art on a standard stream clustering dataset of English documents.

\end{abstract}

\section{Introduction}

Human presentation and understanding of news articles is almost never isolated. Seminal real-world events spawn a chain of strongly correlated news articles that form a news story over time. Given the abundance of online news sources, the consumption of news in the context of the stories they belong to is challenging. Unless people are able to scour the many news sources multiple times a day, major events of interest can be missed as they occur. The real-time monitoring of news, segregating articles into their corresponding stories, thus enables people to follow news stories over time.

This goal of identifying and tracking topics from a news stream was first introduced in the Topic Detection and Tracking (TDT) task \cite{allan1998topic}. Topics in the news stream setting usually correspond to real-world events, while news articles may also be categorized thematically into sports, politics, etc. We focus on the task of clustering news on the basis of event-based story chains. We make a distinction between our definition of an \textit{event topic}, which follows TDT and refers to large-scale real-world events, and the fine-grained events used in trigger-based event detection \cite{ahn-2006-stages}. Given the non-parametric nature of our task (the number of events is not known beforehand and evolves over time), the two primary approaches have been topic modeling using Hierarchical Dirichlet Processes (HDPs) \cite{teh2005sharing, beykikhoshk2018discovering} and Stream Clustering \cite{macqueen1967, laban-hearst-2017-newslens, miranda-etal-2018-multilingual}. While HDPs use word distributions within documents to infer topics, stream clustering models use representation strategies to encode and cluster documents. Contemporary models have adopted stream clustering using TF-IDF weighted bag of words representations to achieve state-of-the-art results \cite{DBLP:conf/ecir/StaykovskiBMN19}.

In this paper, we present a model for event topic detection and tracking from news streams that leverages a combination of dense and sparse document representations. Our dense representations are obtained from BERT models \cite{devlin-etal-2019-bert} fine-tuned using the triplet network architecture \cite{hoffer2015deep} on the event similarity task, which we describe in Section \ref{sec:methodology}. We also use an adaptation of the triplet loss to learn a Support Vector Machine (SVM) \cite{boser1992} based document-cluster similarity model and handle the non-parametric cluster creation using a shallow neural network. We empirically show consistent improvement in clustering performance across many clustering metrics and significantly less cluster fragmentation. 

The main contributions of this paper are:
\begin{itemize}
    \item We present a novel technique for event-driven news stream clustering, which, to the best of our knowledge, is the first attempt of using contextual representations for this task.
    \item We investigate the impact of BERT's fine-tuning objective on clustering performance and show that tuning on the event similarity task using triplet loss improves the effectiveness of embeddings for clustering.
    \item We demonstrate the importance of adding external knowledge to contextual embeddings for clustering by introducing entity awareness to BERT. Contrary to a previous claim \cite{DBLP:conf/ecir/StaykovskiBMN19}, we empirically show that dense embeddings improve clustering performance when augmented with task-pertinent fine-tuning, external knowledge and the conjunction of sparse and temporal representations.
    \item We analyze the problem of cluster fragmentation and show that it is not captured well by the metrics reported in the literature. We propose an additional metric that captures fragmentation better and report results on both.
\end{itemize}

\section{Related Work}

In this section, we introduce the TDT task, prior work on tracking events from news streams and a few related parametric variants of the TDT task.

The goal of the TDT task is to organize a collection of news articles into groups called topics. Topics are defined as sets of highly correlated news articles that are related to some seminal real-world event. This is a narrower definition than the general notion of a topic which could include subjects (like \textit{New York City}) as well. TDT defines an event to be represented by a triple $<$location, time, people involved$>$, which spawns a series of news articles over time. We are interested in all  five sub-tasks of TDT - story segmentation, first story detection, cluster detection, tracking and story link detection - though we do not explicitly tackle these sub-problems individually.

After the initial work on the TDT corpora, interest in news stream clustering was rekindled by the news tracking system \textit{NewsLens} \cite{laban-hearst-2017-newslens}. NewsLens tackled the problem in multiple stages: (1) document representation through keyphrase extraction; (2) non-parametric batch clustering using the Louvian algorithm \cite{blondel2008fast}; and (3) linking of clusters across batches. \citet{DBLP:conf/ecir/StaykovskiBMN19} presented a modified version of this model, using TF-IDF bag of words document representations instead of keywords. They also compared the relative performance of sparse TF-IDF bag of words and dense doc2vec \cite{10.5555/3044805.3045025} representations and showed that the latter performs worse, both individually and in unison with sparse representations. \citet{linger2020batch} extended this batch clustering idea to the multilingual setting by incorporating a Siamese Multilingual-DistilBERT \cite{Sanh2019DistilBERTAD} model to link clusters across languages.

In contrast to the batch-clustering approach, \citet{miranda-etal-2018-multilingual} adopt an online clustering paradigm, where streaming documents are compared against existing clusters to find the best match or to create a new cluster. We adopt this stream clustering approach as it is robust to temporal density variations in the news stream. Batch clustering models tune a batch size hyper-parameter that is both training corpus dependent and might not be able to adjust to temporal variations in stream density. In their model, they also use a pipeline architecture, having separate models for document-cluster similarity computation and cluster creation. Similarity between a document and cluster is computed along multiple document representations and then aggregated using a Rank-SVM model \cite{10.1145/775047.775067}. The decision to merge a document with a cluster or create a new cluster is taken by an SVM classifier. Our model also follows this architecture, but critically adds dense document representations, an SVM trained on the adapted triplet loss for aggregating document-cluster similarities and a shallow neural network for cluster creation.

News event tracking has also been framed as a non-parametric topic modeling problem \cite{zhou-etal-2015-unsupervised} and HDPs that share parameters across temporal batches have been used for this task \cite{beykikhoshk2018discovering}. Dense document representations have been shown to be useful in the parametric variant of our problem, with neural LDA \cite{dieng2019topic, keya2019neural, dieng2019dynamic, bianchi2020pretraining}, temporal topic evolution models \cite{pmlr-v70-zaheer17a, gupta-etal-2018-deep-temporal, 10.1145/3289600.3291036, Brochier_2020} and embedding space clustering \cite{ICWSM1817856, sia2020tired} being some prominent approaches in the literature.

\section{Methodology}
\label{sec:methodology}

\begin{figure*}[t]
  \includegraphics[width=\textwidth]{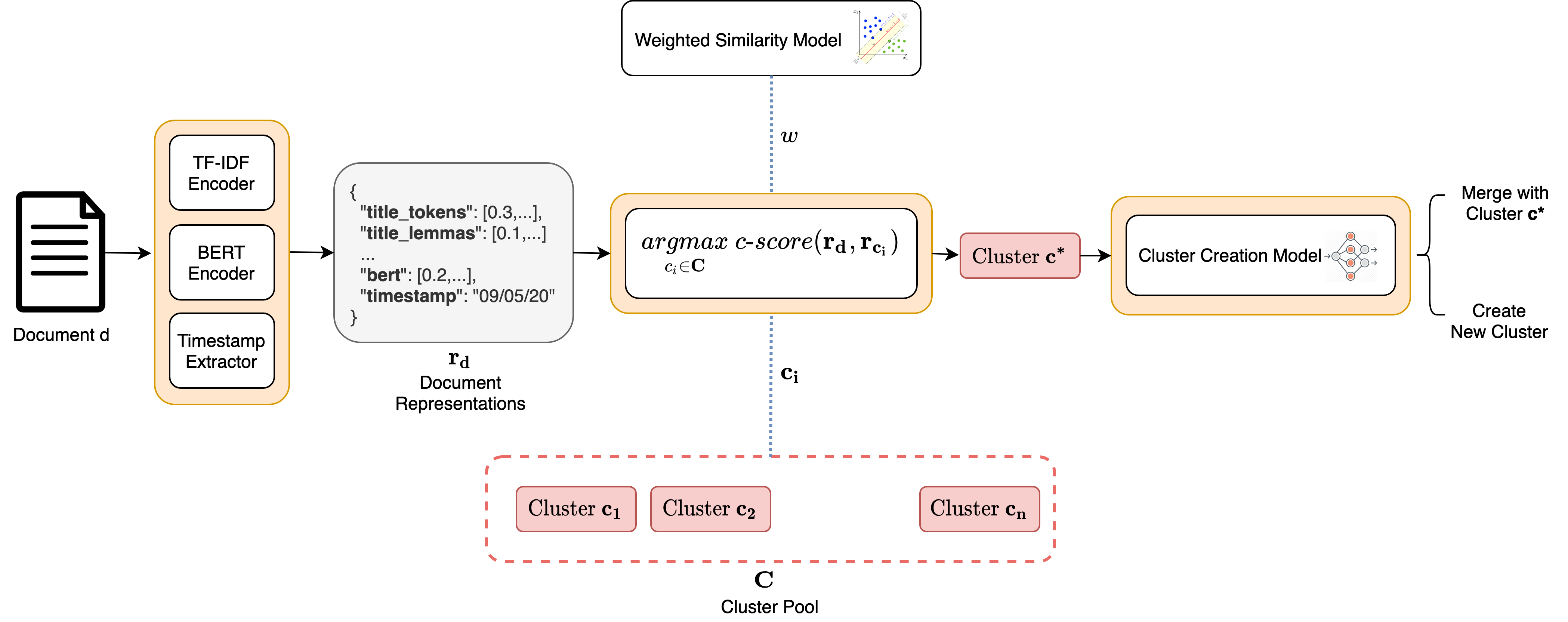}
  \caption{The architecture of the news stream clustering model, showing the clustering process for a single document in the news stream. At the end of the clustering process for each document, the cluster pool is updated based on the output from the cluster creation model, either by adding document $d$ to cluster $c^*$ or by creating a new cluster with the document.}
  \label{image:cluster_model_architecture}
\end{figure*}

Our clustering model is a variant of the streaming K-means algorithm \cite{macqueen1967} with two key differences: (1) we compute the similarity between documents and clusters along a set of representations instead of a single vector representation; and (2) we decide the cluster membership using the output of a neural classifier, a learned model, instead of a static tuned threshold.

At any point in time $t$, let $n$ be the number of clusters the model has created thus far, called the cluster pool. Given a continuous stream of news documents, the goal of the model is to decide the cluster membership (if any) for each input document.
In our task, we assume that each document belongs to a single event, represented by a cluster. The architecture of the model, as shown in Figure \ref{image:cluster_model_architecture}, consists of three main components : (1) document representations, (2) document-cluster similarity computation using a weighted similarity model and (3) cluster creation model. In what follows, we describe each of these components.

\subsection{Document Representations}
\label{ssec:document-representations}

Documents in the news stream have a set of representations, as shown in Figure \ref{image:cluster_model_architecture}, where each representation is one of the following types - sparse TF-IDF, dense embedding or temporal. We describe below these representation types and how clusters, which are created by our model, build representations from their assigned documents.

\subsubsection{TF-IDF Representation}

Separate TF-IDF models that are trained only on the tokens, lemmas and entities in a corpus are used to encode documents separately. For every document in the news stream, its title, body and title+body are each encoded into separate bags of tokens, lemmas and entities, creating nine sparse bag of word representations per document.

\subsubsection{Dense Embedding Representation}

Dense document representations are obtained by embedding the body of documents using BERT, with pre-trained BERT (P-BERT) without any fine-tuning as our baseline embedding model. 
In order to improve the effectiveness of contextual embeddings for our clustering task, we experiment with enhancements along two dimensions: (1) the fine-tuning objective, and (2) the provision of external knowledge. We train separate BERT models for (1) and (2) and use them to encode documents.

To evaluate the impact of the fine-tuning objective, we fine-tune BERT models on two different tasks - event classification (C-BERT) and event similarity (S-BERT). We also evaluate the impact of external knowledge on the embeddings through an entity-aware BERT architecture, which may be paired with either of the fine-tuning objectives.

\paragraph{Fine-tuning on Event Classification}
\label{para:fine-tuning-event-classification}

The goal of this fine-tuning is to tune the CLS token\footnote{\scriptsize The CLS token, introduced in \cite{devlin-etal-2019-bert}, is a special token added to the beginning of every document before being embedded by BERT} embedding such that it encodes information about the event that a document corresponds to. A dense and softmax layer are stacked on top of the CLS token embedding to classify a document into one of the events in the output space.

\paragraph{Fine-tuning on Event Similarity}
\label{para:fine-tuning-event-similarity}

Fine-tuning on the task of event classification constrains the embedding of documents corresponding to different events to be non-linearly separable. Semantics about events can be better captured if the vector similarity between document embeddings encode whether they are from the same event or not. 

For this, we adapt the triplet network architecture \cite{hoffer2015deep} and fine-tune on the task of event similarity. Triplet BERT networks were introduced for the semantic text similarity  (STS) task \cite{reimers-gurevych-2019-sentence}, where the vector similarity between sentence embeddings was tuned to reflect the semantic similarity between them. We formulate the event similarity task, where the term “similarity” refers to whether two documents are from the same event cluster or not. In our task, documents from the same event are similar (with similarity = 1), while those from different events are dissimilar (with similarity = 0). Given the embeddings of an anchor document $d_a$, a positive document $d_p$ (from the same event as the anchor) and a negative document $d_n$ (from a different event), triplet loss is computed as
\begin{equation} \label{eqn:triplet}
    l_{triplet} = sim(d_a, d_n) - sim(d_a, d_p) + m
\end{equation}
where $sim$ is the cosine similarity function and $m$ is the hyper-parameter margin. 


\paragraph{Providing External Entity Knowledge}
\label{para:providing-external-entity-knowledge}

\begin{figure}
    \includegraphics[width=\linewidth]{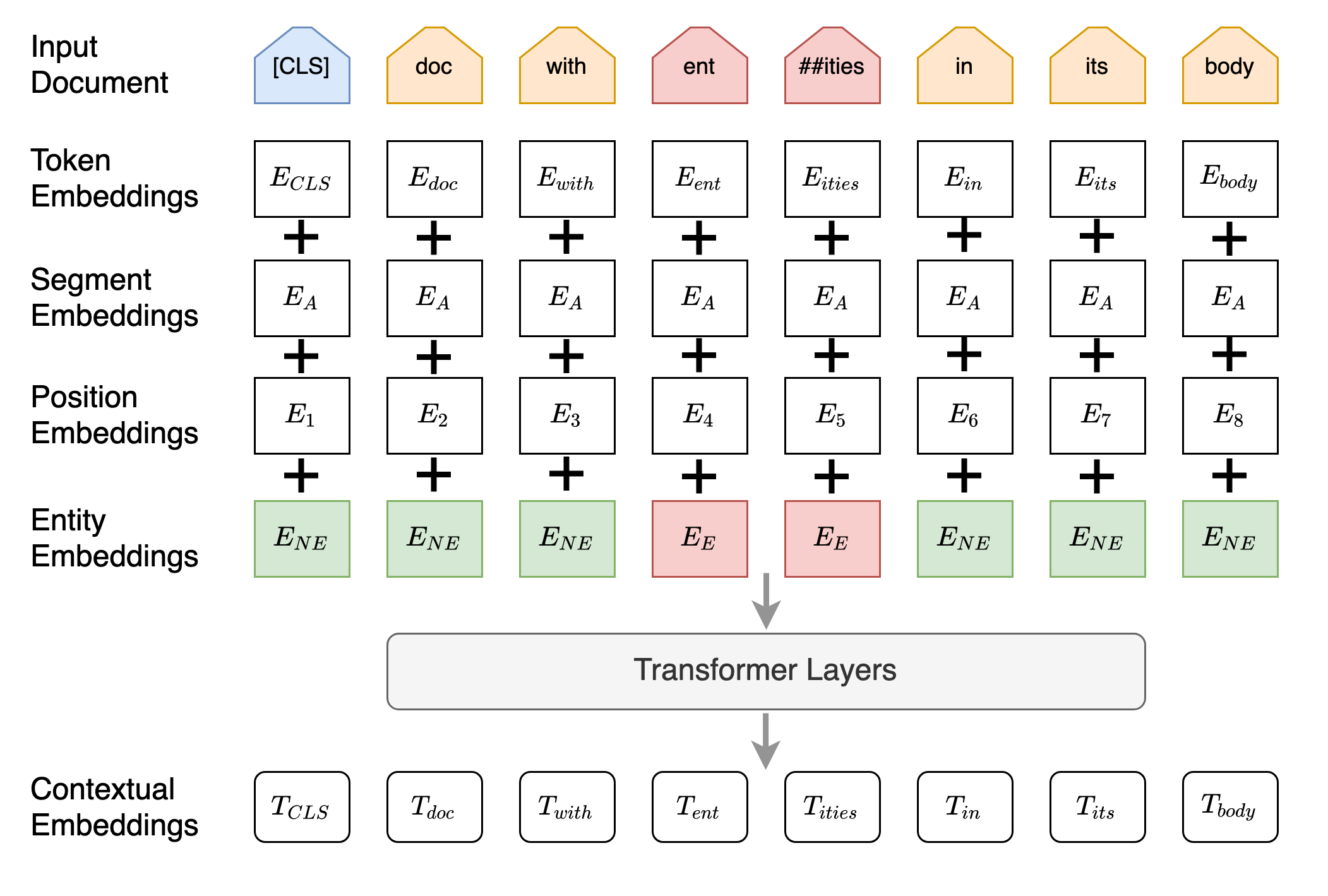}
    \caption{Entity-aware BERT model, with the additional entity presence ($E_{E}$) and absence ($E_{NE}$) embeddings}
    \label{image:entity-bert}
\end{figure}

In line with TDT's definition, entities are central to events and thus need to be highlighted in document representations for our clustering task. We follow \citet{logeswaran-etal-2019-zero} to introduce entity awareness to BERT by leveraging knowledge from an external NER system. Apart from token, position and token type embeddings, we also add an entity presence-absence embedding for each token depending on whether it corresponds to an entity or not. The entity aware BERT model architecture is shown in Figure \ref{image:entity-bert}. 
This enhanced entity-aware model can then be coupled with the event similarity (E-S-BERT) objective for fine-tuning.

\subsubsection{Temporal Representation}

Documents are also represented with the timestamp of publication. Unlike TF-IDF and dense embeddings, which are vector valued representations, the temporal representation of a document is just a single value (e.g "05-09-2020") which has an associated subtraction operation. The difference between two timestamps is defined as the number of intervening days between them. Section \ref{ssec:weighted-sim-model} describes how these timestamps are used for clustering.

\subsubsection{Cluster Representation}

Since clusters are created and updated by our model, their representations need to be generated dynamically from the documents assigned to them. While documents in the news stream have a set of 11 representations (9 TF-IDF, dense embeddings and timestamp), clusters have two additional time\-stamp representations. Cluster representations are derived from documents in the cluster through aggregation. While dense embedding and sparse TF-IDF representations of a cluster are aggregated using mean pooling, clusters have three timestamp representations corresponding to different aggregation strategies - min, max and mean pooling.


\subsection{Weighted Similarity Model}
\label{ssec:weighted-sim-model}

Once documents are encoded by a set of representations, they are compared to the clusters in the cluster pool to find the most compatible cluster. The similarity between a document and a cluster is computed along each representation separately and is then aggregated into a single compatibility score ($c\text{-}score$). While similarity along contextual embeddings and TF-IDF bag representations is computed using cosine similarity (as shown in Equation \ref{eqn:cosine}), timestamp similarity is computed using the Gaussian similarity function introduced in \citet{miranda-etal-2018-multilingual} (as shown in Equation \ref{eqn:gaussian}).

\begin{figure*}
    \centering
    \includegraphics[width=0.75\linewidth]{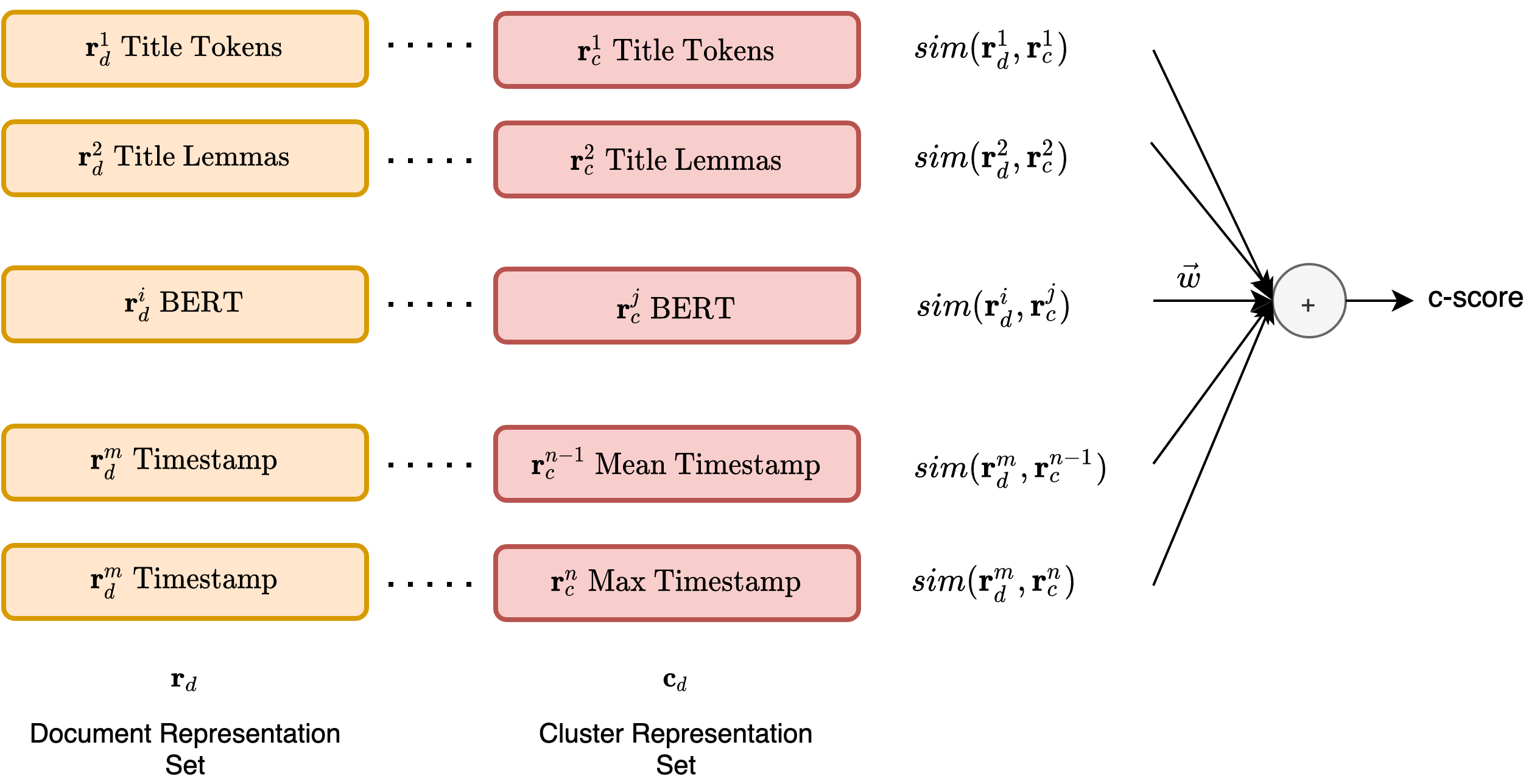}
    \caption{Computation of c-score: (a) similarities are computed for each representation individually using the appropriate similarity function (cosine or Gaussian); (b) subsequently, the computed similarities are aggregated into a single c-score value using the weights of the weighted similarity model ($w$)}
    \label{image:c-score}
\end{figure*}

Let $\mathbf{r_d}^v$ and $\mathbf{r_c}^v$ denote a dense or sparse vector representation of a document $d$ and cluster $c$ respectively. Let $\mathbf{r_d}^t$ and $\mathbf{r_c}^t$ denote their timestamp representations. Let $(i, j)$ correspond to a pair of document-cluster representations of the same type (as defined in Section \ref{ssec:document-representations}). Document-cluster similarity is computed along each representation and aggregated using a weighted summation as
\begin{align}
    sim(\mathbf{r_d}, \mathbf{r_c}) &= \{ sim(\mathbf{r_d}^i, \mathbf{r_c}^j)\ \forall\ (i, j)\} \nonumber \\
    sim(\mathbf{r_d}^v, \mathbf{r_c}^v) &= \frac{\mathbf{r_d}^v \cdot \mathbf{r_c}^v}{| \mathbf{r_d}^v| |\mathbf{r_c}^v|} \label{eqn:cosine} \\
    sim(\mathbf{r_d}^t, \mathbf{r_c}^t) &= e^{-\frac{((\mathbf{r_d}^t - \mathbf{r_c}^t) - \mu)}{2 \sigma^2}} \label{eqn:gaussian} \\
    c\text{-}score(\mathbf{r_d}, \mathbf{r_c}) &= \sum_{(i, j)} w_j \cdot sim(\mathbf{r_d}^i, \mathbf{r_c}^j) \nonumber
\end{align}
where $\mu$ and $\sigma$ are tuned hyper-parameters of the temporal similarity function. It is noted here that since clusters have two additional timestamp representations, all three timestamp similarities are computed using the single document timestamp representation, as illustrated in Figure \ref{image:c-score}.

The dataset does not contain annotation for the degree of membership between a document and cluster and thus, the weights for combining the representation similarities can't be learned directly. To circumvent this issue, we train a linear model on a relevant task so that the trained weights can then be adapted to compute the compatibility score. 

In our model, we train a linear model on a novel adaptation of the event similarity triplet loss used to train the S-BERT model. 
The triplet loss, as defined in Equation \ref{eqn:triplet}, can be adapted to a linear classifier if similarity has a related notion with regards to the classifier. SVM is an appropriate model since the degree of compatibility between a point $x$ and a class $c$ is given by the distance of the point from the class' decision hyperplane $w_c$. This distance, computed as $w_c \cdot x + b$, can thus be used as the similarity metric to adapt the triplet loss. 

In our case, the inputs to the SVM model are vectors of document-cluster similarities along the set of representations $sim(\mathbf{r_d}, \mathbf{r_c})$. The adapted SVM-triplet loss is thus computed as shown below. Since we want to minimize this loss, we analyze its point of minima.
\begin{align*}
    l_{svm-triplet} &= w \cdot sim(\mathbf{r_a}, \mathbf{r_n}) - w \cdot sim(\mathbf{r_a}, \mathbf{r_p}) \\
    &\qquad + m \\
    l_{svm-triplet} &= 0 \\
    \implies m &= w \cdot (sim(\mathbf{r_a}, \mathbf{r_p}) - sim(\mathbf{r_a}, \mathbf{r_n}))
\end{align*}

The adapted triplet loss can thus be modeled as a classification task with inputs $(sim(\mathbf{r_a}, \mathbf{r_p}) - sim(\mathbf{r_a}, \mathbf{r_n}))$ and the outputs $m$. For mathematical convenience, we set $m=1$ without loss of generality. In this manner, we transform the event similarity triplet loss objective into a classification objective to train an SVM model. The novelty of this supervision is that we focus on learning useful weights and not a useful classifier. The learned weights, which minimize the event similarity triplet loss, are utilized for document-cluster c-score computation. During the clustering process, a document $d$ is compared against all the clusters in the pool $\mathbf{C}$ to determine the most compatible cluster $c^*$ as
\begin{equation*}
    c^* = \argmax_{c\ \in\ \mathbf{C}}\ c\text{-}score(\mathbf{r_d}, \mathbf{r_c})
\end{equation*}

\subsection{Cluster Creation Model}

Since our clustering problem is non-parametric, each document in the stream could potentially be the start of a new event cluster. Thus, the most compatible cluster $c^*$ might not actually be the cluster that the document corresponds to. Given a document and its most compatible cluster, the cluster creation model decides whether or not a new cluster is to be created. For this, we employ a shallow neural network which takes document-cluster similarities along the set of representations as input and decides if a new cluster should be created. Since the dimensionality of the input space for the network is small, we use a shallow network to prevent overfitting.

\section{Experiments and Results}
\label{sec:results}

\subsection{Data}

To train and evaluate our clustering models, we use the standard multilingual news stream clustering dataset \cite{miranda-etal-2018-multilingual}, which contains articles from English, French and German. For our clustering task, we only use the English subset of the corpus, which consists of 20,959 articles. Articles are annotated with language, timestamp and the event cluster to which they belong, in addition to their title and body text. We use the same training and evaluation split provided by \citet{miranda-etal-2018-multilingual} and use the training set to fine-tune the parameters of the clustering model. The training and evaluation sets are \textit{temporally disjoint} to ensure that the clustering models are tuned independent of the events seen during training.

\subsection{Experimental Setup}

We train our model pipeline in a sequence where each component model is supplied with the output from the component trained before in the sequence.
For instance, the cluster creation model is trained using the embeddings from the fine-tuned BERT model and by selecting the most compatible cluster determined by the trained weighted similarity model. We experiment with multiple document representation sets, training all the component models each time and evaluating the entire clustering model on the test set.

We use the TF-IDF weights provided in the \citet{miranda-etal-2018-multilingual} corpus to ensure fair comparison with prior work. For training the event similarity BERT model (S-BERT), triplets are generated for each document using the batch-hard regime \cite{alex2017defense} by picking the hardest positive and negative examples from its mini-batch\footnote{We use the batch-hard implementation provided by \citet{reimers-gurevych-2019-sentence} at \url{https://github.com/UKPLab/sentence-transformers}}. We train the S-BERT model for 2 epochs using a batch size of 32, with 10\% of the training data being used for linear warmup. We use Adam optimizer with learning rate $2e^{-5}$ and epsilon $1e^{-6}$. Document embeddings are obtained by mean pooling across all its tokens. For NER, we use the medium English model provided by spaCy \cite{spacy2}.

Training instances for the weighted similarity and cluster creation models are generated by simulating the stream clustering process on the training set and assigning each document to its true event cluster. For the weighted similarity model, we generate triples of $<$document, true cluster, sampled negative cluster$>$ and convert them into SVM training instances as mentioned in Section \ref{ssec:weighted-sim-model}. Since all the training instances have the same label $m$, half the training set is negated to balance the dataset. 

To generate training samples for the cluster creation model, the most compatible cluster is determined using the trained weighted similarity model for each document. A sample is then generated with input as the document-cluster similarities and output as 0 or 1 depending on whether the true cluster for that document is in the cluster pool or not. The dataset contains over 12k documents but only 593 clusters, entailing that the fraction of training samples where a new cluster is created is only 5\%, making the dataset extremely biased. To mitigate this issue, we use the SVMSMOTE algorithm \cite{10.1504/IJKESDP.2011.039875} to oversample the minority class and make the classes equal in size. For cluster creation, we train a shallow single layer neural network with two nodes using the L-BFGS solver \cite{10.2307/2006193}. The weighted similarity and cluster creation models are trained using 5-fold cross validation to tune hyper-parameters and then on the entire training set using the best settings.

The clustering output is  evaluated by comparing against the ground truth clusters. We report results on the B-Cubed metrics \cite{bagga1998algorithms} in Table \ref{table:clustering-comparison} to compare against prior work.

\begin{table*}
\resizebox{\textwidth}{!}{
\small
\begin{tabular}{lcccc}
\hline
\multicolumn{1}{c}{\multirow{2}{*}{\textbf{Model}}} & \multirow{2}{*}{\textbf{\begin{tabular}[c]{@{}c@{}}Clusters Count\\ (True Count - 222)\end{tabular}}} & \multicolumn{3}{c}{\textbf{B-Cubed Metrics}}              \\
\multicolumn{1}{c}{}                                &                                                                                                       & \textbf{Precision} & \textbf{Recall} & \textbf{$F_1$ Score} \\
\hline
\citet{laban-hearst-2017-newslens}                                            & 873                                                                                                   & 94.37              & 85.58           & 89.76              \\
\citet{miranda-etal-2018-multilingual}                                      & 326                                                                                                   & 94.27              & 90.25           & 92.36              \\
\citet{DBLP:conf/ecir/StaykovskiBMN19}                                   & 484                                                                                                   & 95.16              & 93.66           & 94.41              \\
\citet{linger2020batch}                                        & 298                                                                                                   & 94.19              & 93.55           & 93.86              \\
\hline
Ours - TF-IDF                                       & 530                                                                                                   & 93.50              & 80.23           & 86.36              \\
Ours - TF-IDF (out-of-order)                                       & 413                                                                                                   & 90.57              & 87.51           & 89.01              \\
Ours - TF-IDF + Time                                & 222                                                                                                   & 87.57              & 96.27           & 91.72              \\
\hline
Ours - E-S-BERT                                     & 452                                                                                                   & 79.76              & 60.77           & 68.98              \\
Ours - E-S-BERT + Time                              & 471                                                                                                   & 92.70               & 74.69           & 82.73              \\
\hline
Ours - TF-IDF + P-BERT + Time                       & 196                                                                                                   & 83.12              & \textbf{97.26}  & 89.63              \\
Ours - TF-IDF + C-BERT + Time                       & 321                                                                                                   & 83.10               & 91.33           & 87.03              \\
Ours - TF-IDF + S-BERT + Time                       & 247                                                                                                   & 88.30               & 96.10           & 92.04

 \\
\hline
Ours - TF-IDF + E-S-BERT            &433                                                                                                    &89.40      & 86.99           & 88.18    \\
Ours - TF-IDF + E-S-BERT (out-of-order)     &384                                                                                            &91.15      &88.60       &89.86     \\
Ours - TF-IDF + E-S-BERT + Time                     & 276                                                                                                   & \textbf{94.28}     & 95.25           & \textbf{94.76}     \\
\hline
\end{tabular}
}
\caption{Results of clustering performance for different document representation strategies as compared against contemporary models. P-BERT refers to pre-trained BERT; C-BERT refers to BERT fine-tuned on event classification\; S-BERT refers to BERT fine-tuned using triplet loss on event similarity; E-S-BERT  refers to entity aware BERT fine-tuned on event similarity.}
\label{table:clustering-comparison}
\end{table*}

\subsection{Results}

\paragraph{TF-IDF sets a tough baseline:}

Prior work has shown that sparse TF-IDF bag representations achieve competitive performance~\cite{laban-hearst-2017-newslens, miranda-etal-2018-multilingual} and our experiments validate this observation. The clustering model that uses only sparse TF-IDF bags to represent documents achieves a very high score of 86.8\% B-Cubed $F_1$ score, as shown in Table~\ref{table:clustering-comparison}. If TF-IDF bags are used in combination with timestamps, then the performance further increases to 91.7\%, setting a tough baseline to beat.

\paragraph{Contextual embeddings, by themselves, achieve sub-par clustering performance:}

In line with prior work, we observe that dense document embeddings, both when used as the sole representation and in conjunction with timestamps, are unable to match the clustering performance of TF-IDF bags. It can be seen in Table \ref{table:clustering-comparison} that even our best BERT model (entity aware BERT trained on event similarity) only achieves an F1 score of 69\% individually and 82.7\% when combined with timestamp representations. These scores are 17.8\% and 9\% lower than their corresponding TF-IDF counterparts. BERT embeddings are richer representations that encode linguistic information including syntax and semantics through its pre-training. Thus, the model is unable to distinguish between events at the desired granularity and ends up clustering together topically related events (for instance, two different events related to soccer). 


\paragraph{Fine-tuning objective impacts the effectiveness of embeddings for clustering:}

In most NLP tasks, fine-tuning contextual embeddings on a related pertinent objective is beneficial, we observe that the choice of fine-tuning objective is critical to the task performance. While the baseline pre-trained P-BERT model achieves a clustering score of 89.6\% when used in conjunction with TF-IDF and timestamp representations (TF-IDF + P-BERT + Time), fine-tuning embeddings on event classification (TF-IDF + C-BERT + Time) drops the performance to 87\%. This drop in performance can be attributed to the following reasons. Firstly, the large output space (593 events) and small dataset size (12k documents) make it hard for the model to learn effectively during fine-tuning. In addition to this, the classification objective requires that the embeddings of documents from different events be non-linearly separable. But this is not directly compatible with how the embeddings are used by the weighted similarity model, which is to compute cosine similarity. This discordance entails that the fine-tuning process degrades the clustering performance. The event similarity triplet loss is a more suitable fine-tuning objective and it is observed that fine-tuning BERT on this objective (TF-IDF + S-BERT + Time) results in a better clustering performance of 92.04\%.

\begin{table*}[h]
\centering
\small
\begin{tabular}{lccr}
\hline
\multicolumn{1}{c}{\textbf{Metric}}                      & \textbf{TF-IDF + E-S-BERT + Time} & \textbf{Miranda} & \textbf{Gain}           \\
\hline
B-Cubed                           & 94.76                                    & 92.36            &           2.40$^\dagger$                     \\
CEAF-e        & 76.93                                    & 69.57            &               7.36$^\dagger$                 \\
CEAF-m                            & 93.31                                    & 90.19            &        3.12$^\dagger$                        \\
MUC                            & 99.30                                     & 98.88            &           0.42$^\ddagger$                     \\
BLANC                            & 98.13                                    & 96.93            &              1.20$^\mathsection$                  \\
V Measure                            & 97.98                                    & 97.01            &    0.97$^\dagger$                            \\
Adjusted Rand Score                  & 96.26                                    & 93.87            &    2.39$^\mathsection$                            \\
Adjusted Mutual Information                  & 97.99                                    & 97.02            &    2.97$^\mathsection$                            \\
Fowlkes Mallows Score                & 96.38                                    & 94.11            & 2.27$^\mathsection$          \\
\hline
\end{tabular}
\caption{Results of clustering performance across different evaluation metrics. For each metric computed using precision, recall and F-1 scores, only the F-1 scores are reported. Statistically significant gains, with $p <<< 0.001$ are denoted by $^\dagger$ and $p < 0.01$ by  $^\ddagger$. Gains denoted by $\mathsection$ are not evaluated for significance, in line with literature.}
\label{table:metrics-comparison}
\end{table*}

\paragraph{External entity knowledge makes embeddings more effective for clustering:}

The introduction of external knowledge through the entity aware BERT architecture significantly improves the clustering performance of the model. It can be seen in Table \ref{table:clustering-comparison} that introducing entity awareness and training on the event similarity task (TF-IDF + E-S-BERT + Time) results in a clustering score of 94.76\%\footnote{The mean and standard deviation of the precision, recall and F-1 scores over five independent training and evaluations of our model are $94.64 \pm 0.28$, $94.72 \pm 1.33$ and $94.75 \pm 0.59$.}, achieving a new state-of-the-art on the dataset\footnote{We observe similar results on the TDT Pilot dataset~\cite{allan1998topic}, as shown in Section~\ref{ss:restdt}}. 
The results are statistically significant 
and $p$ values from a paired student's t-test are reported in Table~\ref{table:metrics-comparison}. This is  almost 3 points better than the corresponding model without entity awareness, which highlights the importance of this external knowledge. When given external knowledge from an NER system, the BERT model, like sparse TF-IDF representations, is able to draw attention to entities and highlight them in the document embeddings. It is observed that the model learns to project entities and non-entities in mutually orthogonal directions and thereby adds emphasis to entities.

In our experiments, we observe an increase of almost 1 point in $F_1$ score by considering only a subset of the OntoNotes corpus~\cite{w13} labels \footnote{Our entity label subset consists PERSON, NORP, FAC, ORG, GPE, LOC, PRODUCT, EVENT, WORK OF ART, LAW and LANGUAGE.}. Ignoring entity classes like ORDINAL and CARDINAL helps as they don't provide useful information for our clustering task. The scores reported in Table \ref{table:clustering-comparison} correspond to entity-aware models trained on this label subset. We also experimented with separate embeddings for each entity type instead of the binary entity presence-absence embeddings and observed that it degrades $F_1$ score by more than 2 points.

\paragraph{Ablating time and non-streaming input:}
When we ablate timestamp from the representation (rows that are not marked with ``Time" in Table~\ref{table:clustering-comparison}) and then stream documents in random order
(rows marked with (``out- of-order") in Table~\ref{table:clustering-comparison}), 
the number of clusters increase over when accounting for time.  When ablating time, we also observe that supplying documents 
in random order
produces fewer clusters and better b-cubed  $F_1$ scores. We observe examples of clusters that are incorrectly merged in the absence of temporal information (in the out-of-order setting). See Appendix for actual examples from our output.

\paragraph{Cluster fragmentation is not captured well by B-Cubed metrics}
The improvements our model makes can be seen clearly by observing the number of clusters created by the model. While the previous state-of-the-art model produced 484 clusters, ours produces only 276\footnote{The mean and standard deviation of the cluster count over five independent training and evaluations of our model are $312 \pm 27$.}, which is closer to the true cluster count of 222. Our model produces far less cluster fragmentation, resulting in a 79.4\% reduction in the number of erroneous additional clusters created. We argue that this is an important improvement that is not well captured by the small increase in B-Cubed metrics. 

While B-Cubed $F_1$ score is the standard metric reported in the literature, it is an article-level metric which gives more importance to large clusters. This entails that B-Cubed metrics heavily penalize the model's output for making mistakes on large clusters while mistakes on smaller clusters can fall through without incurring much penalty. In our experiments, we observed that this property of the metric prevents it from capturing cluster fragmentation errors on smaller events. 
In the news stream clustering setting, small events may correspond to recent salient events and thus, we want our metric to be agnostic to the size of the clusters.

We thus use an additional metric that weights every cluster equally - CEAF-e \cite{luo-2005-coreference}. The CEAF-e metric creates a one-to-one mapping between the clustering output and gold clusters using the Kuhn-Munkres algorithm. The similarity between a gold cluster $G$ and an output cluster $O$ is computed as the fraction of articles that are common to the clusters. 
Once the clusters are aligned, precision and recall are computed using the aligned pairs of clusters. This ensures that unaligned clusters contribute to a penalty in the score and cluster fragmentation and coalescing is captured by the metric.

In order to ensure that our model's better performance is metric-agnostic, we also empirically evaluated our clustering model against prior work using several clustering metrics, the results of which are presented in Table \ref{table:metrics-comparison}. 
For this, we compare with \citet{miranda-etal-2018-multilingual} since their results are readily replicable.
It can be observed that our model consistently achieves better performance across most metrics and is thus robust to the metric idiosyncrasies. Our model achieves an improvement of 7.36 points on the CEAF-e metric, which shows that our clustering model performs better than contemporary models on smaller clusters as well. 


\subsection{Results on TDT}
\label{ss:restdt}
To validate the robustness of our clustering model, we evaluate it on the TDT Pilot corpus \cite{allan1998topic}. The TDT Pilot corpus consists of a set of newswire and broadcast news transcripts that span the period from July 1, 1994 to June 30, 1995. Out of the 16,000 documents collected, 1,382 are annotated to be relevant to one of 25 events during that period. Unlike the \citet{miranda-etal-2018-multilingual} corpus, TDT Pilot does not have the article titles. We, therefore, train all the components of our ensemble architecture using only the document body text. The TDT corpus does not provide pre-trained TF-IDF weights, so we train the weights on the entire corpus as a pre-processing step. Unlike Miranda, the TDT Corpus
also lacks standard train and test splits. We create our own splits across 25 events. The splits are described and listed in the Appendix.

In line with our observations on the \citet{miranda-etal-2018-multilingual} corpus, we observe  similar results on the TDT corpus. 
We achieve the best result on this corpus on a model with TF-IDF representations combined with 
temporal representations, BERT entity-aware representations fine-tuned on the event similarity task.
The best result has a b-cubed precision of \textbf{81.62}, b-cubed recall of \textbf{95.89} and a b-cubed \textbf{$F_1$ of 88.18}. We generate \textbf{12 clusters} which matches the number of clusters in the ground truth.

We show that even in a cross-corpus setting, dense contextual embeddings, when augmented with pertinent fine-tuning, external knowledge and the conjunction of sparse and temporal representations, are a potent representation strategy for event topic clustering.

\section{Conclusion}

In this paper, we present a novel news stream clustering algorithm that uses a combination of sparse and dense vector representations. We show that while dense embeddings by themselves do not achieve the best clustering results, enhancements like entity awareness and event similarity fine-tuning make them effective in conjunction with sparse and temporal representations. 
Our model achieves new state-of-the-art results on the \citet{miranda-etal-2018-multilingual} dataset. 
We also analyze the problem of cluster fragmentation noting that our approach is able to produce a similar number of clusters as in the test set, in contrast to prior work which produces far too many clusters. We note issues with 
the B-Cubed metrics and we complement our results using CEAF-e as an additional metric for our clustering task. In addtion, we provide a comprehensive empirical evaluation across many metrics to show the robustness of our model to metric idiosyncrasies.

\section*{Acknowledgments}
We thank Chao Zhao, Heng Ji, Rishita Anubhai and Graham Horwood for valuable discussions.

\bibliography{anthology,eacl2021}

\begin{thebibliography}{36}
\expandafter\ifx\csname natexlab\endcsname\relax\def\natexlab#1{#1}\fi

\bibitem[{Ahn(2006)}]{ahn-2006-stages}
David Ahn. 2006.
\newblock \href {https://www.aclweb.org/anthology/W06-0901} {The stages of
  event extraction}.
\newblock In \emph{Proceedings of the Workshop on Annotating and Reasoning
  about Time and Events}, pages 1--8, Sydney, Australia. Association for
  Computational Linguistics.

\bibitem[{Allan et~al.(1998)Allan, Carbonell, Doddington, Yamron, and
  Yang}]{allan1998topic}
James Allan, Jaime~G Carbonell, George Doddington, Jonathan Yamron, and Yiming
  Yang. 1998.
\newblock Topic detection and tracking pilot study final report.
\newblock In \emph{Proceedings of the DARPA Broadcast News Transcription and
  Understanding Workshop}, pages 194--218.

\bibitem[{Bagga and Baldwin(1998)}]{bagga1998algorithms}
Amit Bagga and Breck Baldwin. 1998.
\newblock Algorithms for scoring coreference chains.
\newblock In \emph{The first international conference on language resources and
  evaluation workshop on linguistics coreference}, volume~1, pages 563--566.
  Citeseer.

\bibitem[{Beykikhoshk et~al.(2018)Beykikhoshk, Arandjelovi{\'c}, Phung, and
  Venkatesh}]{beykikhoshk2018discovering}
Adham Beykikhoshk, Ognjen Arandjelovi{\'c}, Dinh Phung, and Svetha Venkatesh.
  2018.
\newblock Discovering topic structures of a temporally evolving document
  corpus.
\newblock \emph{Knowledge and Information Systems}, 55(3):599--632.

\bibitem[{{Bianchi} et~al.(2020){Bianchi}, {Terragni}, and
  {Hovy}}]{bianchi2020pretraining}
Federico {Bianchi}, Silvia {Terragni}, and Dirk {Hovy}. 2020.
\newblock The dynamic embedded topic model.
\newblock \emph{arXiv preprint arXiv:2004.03974}.

\bibitem[{Blondel et~al.(2008)Blondel, Guillaume, Lambiotte, and
  Lefebvre}]{blondel2008fast}
Vincent~D Blondel, Jean-Loup Guillaume, Renaud Lambiotte, and Etienne Lefebvre.
  2008.
\newblock Fast unfolding of communities in large networks.
\newblock \emph{Journal of statistical mechanics: theory and experiment},
  2008(10):P10008.

\bibitem[{Boser et~al.(1992)Boser, Guyon, and Vapnik}]{boser1992}
Bernhard~E. Boser, Isabelle~M. Guyon, and Vladimir~N. Vapnik. 1992.
\newblock \href {https://doi.org/10.1145/130385.130401} {A training algorithm
  for optimal margin classifiers}.
\newblock In \emph{Proceedings of the Fifth Annual Workshop on Computational
  Learning Theory}, COLT '92, page 144–152, New York, NY, USA. Association
  for Computing Machinery.

\bibitem[{Brochier et~al.(2020)Brochier, Guille, and Velcin}]{Brochier_2020}
Robin Brochier, Adrien Guille, and Julien Velcin. 2020.
\newblock \href {https://doi.org/10.1007/978-3-030-45439-5_22} {Inductive
  document network embedding with topic-word attention}.
\newblock \emph{Advances in Information Retrieval}, page 326–340.

\bibitem[{Devlin et~al.(2019)Devlin, Chang, Lee, and
  Toutanova}]{devlin-etal-2019-bert}
Jacob Devlin, Ming-Wei Chang, Kenton Lee, and Kristina Toutanova. 2019.
\newblock \href {https://doi.org/10.18653/v1/N19-1423} {{BERT}: Pre-training of
  deep bidirectional transformers for language understanding}.
\newblock In \emph{Proceedings of the 2019 Conference of the North {A}merican
  Chapter of the Association for Computational Linguistics: Human Language
  Technologies, Volume 1 (Long and Short Papers)}, pages 4171--4186,
  Minneapolis, Minnesota. Association for Computational Linguistics.

\bibitem[{Dieng et~al.(2019{\natexlab{a}})Dieng, Ruiz, and
  Blei}]{dieng2019topic}
Adji~B Dieng, Francisco J~R Ruiz, and David~M Blei. 2019{\natexlab{a}}.
\newblock Topic modeling in embedding spaces.
\newblock \emph{arXiv preprint arXiv:1907.04907}.

\bibitem[{Dieng et~al.(2019{\natexlab{b}})Dieng, Ruiz, and
  Blei}]{dieng2019dynamic}
Adji~B Dieng, Francisco~JR Ruiz, and David~M Blei. 2019{\natexlab{b}}.
\newblock The dynamic embedded topic model.
\newblock \emph{arXiv preprint arXiv:1907.05545}.

\bibitem[{Gupta et~al.(2018)Gupta, Rajaram, Sch{\"u}tze, and
  Andrassy}]{gupta-etal-2018-deep-temporal}
Pankaj Gupta, Subburam Rajaram, Hinrich Sch{\"u}tze, and Bernt Andrassy. 2018.
\newblock \href {https://doi.org/10.18653/v1/N18-1098} {Deep
  temporal-recurrent-replicated-softmax for topical trends over time}.
\newblock In \emph{Proceedings of the 2018 Conference of the North {A}merican
  Chapter of the Association for Computational Linguistics: Human Language
  Technologies, Volume 1 (Long Papers)}, pages 1079--1089, New Orleans,
  Louisiana. Association for Computational Linguistics.

\bibitem[{Hermans et~al.(2017)Hermans, Beyer, and Leibe}]{alex2017defense}
Alexander Hermans, Lucas Beyer, and Bastian Leibe. 2017.
\newblock In defense of the triplet loss for person re-identification.
\newblock \emph{arXiv preprint arXiv:1703.07737}.

\bibitem[{Hoffer and Ailon(2015)}]{hoffer2015deep}
Elad Hoffer and Nir Ailon. 2015.
\newblock Deep metric learning using triplet network.
\newblock In \emph{International Workshop on Similarity-Based Pattern
  Recognition}, pages 84--92. Springer.

\bibitem[{Honnibal and Montani(2017)}]{spacy2}
Matthew Honnibal and Ines Montani. 2017.
\newblock {spaCy 2}: Natural language understanding with {B}loom embeddings,
  convolutional neural networks and incremental parsing.
\newblock To appear.

\bibitem[{Joachims(2002)}]{10.1145/775047.775067}
Thorsten Joachims. 2002.
\newblock \href {https://doi.org/10.1145/775047.775067} {Optimizing search
  engines using clickthrough data}.
\newblock In \emph{Proceedings of the Eighth ACM SIGKDD International
  Conference on Knowledge Discovery and Data Mining}, KDD '02, page 133–142,
  New York, NY, USA. Association for Computing Machinery.

\bibitem[{Keya et~al.(2019)Keya, Papanikolaou, and Foulds}]{keya2019neural}
Kamrun~Naher Keya, Yannis Papanikolaou, and James~R. Foulds. 2019.
\newblock Neural embedding allocation: Distributed representations of topic
  models.
\newblock \emph{arXiv preprint arXiv:1909.04702}.

\bibitem[{Laban and Hearst(2017)}]{laban-hearst-2017-newslens}
Philippe Laban and Marti Hearst. 2017.
\newblock \href {https://doi.org/10.18653/v1/W17-2701} {news{L}ens: building
  and visualizing long-ranging news stories}.
\newblock In \emph{Proceedings of the Events and Stories in the News Workshop},
  pages 1--9, Vancouver, Canada. Association for Computational Linguistics.

\bibitem[{Le and Mikolov(2014)}]{10.5555/3044805.3045025}
Quoc Le and Tomas Mikolov. 2014.
\newblock Distributed representations of sentences and documents.
\newblock In \emph{Proceedings of the 31st International Conference on
  International Conference on Machine Learning - Volume 32}, ICML’14, page
  II–1188–II–1196. JMLR.org.

\bibitem[{Linger and Hajaiej(2020)}]{linger2020batch}
Mathis Linger and Mhamed Hajaiej. 2020.
\newblock Batch clustering for multilingual news streaming.
\newblock \emph{arXiv preprint arXiv:2004.08123}.

\bibitem[{Logeswaran et~al.(2019)Logeswaran, Chang, Lee, Toutanova, Devlin, and
  Lee}]{logeswaran-etal-2019-zero}
Lajanugen Logeswaran, Ming-Wei Chang, Kenton Lee, Kristina Toutanova, Jacob
  Devlin, and Honglak Lee. 2019.
\newblock \href {https://doi.org/10.18653/v1/P19-1335} {Zero-shot entity
  linking by reading entity descriptions}.
\newblock In \emph{Proceedings of the 57th Annual Meeting of the Association
  for Computational Linguistics}, pages 3449--3460, Florence, Italy.
  Association for Computational Linguistics.

\bibitem[{Luo(2005)}]{luo-2005-coreference}
Xiaoqiang Luo. 2005.
\newblock \href {https://www.aclweb.org/anthology/H05-1004} {On coreference
  resolution performance metrics}.
\newblock In \emph{Proceedings of Human Language Technology Conference and
  Conference on Empirical Methods in Natural Language Processing}, pages
  25--32, Vancouver, British Columbia, Canada. Association for Computational
  Linguistics.

\bibitem[{MacQueen(1967)}]{macqueen1967}
J.~MacQueen. 1967.
\newblock \href {https://projecteuclid.org/euclid.bsmsp/1200512992} {Some
  methods for classification and analysis of multivariate observations}.
\newblock In \emph{Proceedings of the Fifth Berkeley Symposium on Mathematical
  Statistics and Probability, Volume 1: Statistics}, pages 281--297, Berkeley,
  Calif. University of California Press.

\bibitem[{Miranda et~al.(2018)Miranda, Znoti{\c{n}}{\v{s}}, Cohen, and
  Barzdins}]{miranda-etal-2018-multilingual}
Sebasti{\~a}o Miranda, Art{\=u}rs Znoti{\c{n}}{\v{s}}, Shay~B. Cohen, and
  Guntis Barzdins. 2018.
\newblock \href {https://doi.org/10.18653/v1/D18-1483} {Multilingual clustering
  of streaming news}.
\newblock In \emph{Proceedings of the 2018 Conference on Empirical Methods in
  Natural Language Processing}, pages 4535--4544, Brussels, Belgium.
  Association for Computational Linguistics.

\bibitem[{Momeni et~al.(2018)Momeni, Karunasekera, Goyal, and
  Lerman}]{ICWSM1817856}
Elaheh Momeni, Shanika Karunasekera, Palash Goyal, and Kristina Lerman. 2018.
\newblock Modeling evolution of topics in large-scale temporal text corpora.
\newblock In \emph{Twelfth International AAAI Conference on Web and Social
  Media}.

\bibitem[{Nguyen et~al.(2011)Nguyen, Cooper, and
  Kamei}]{10.1504/IJKESDP.2011.039875}
Hien~M. Nguyen, Eric~W. Cooper, and Katsuari Kamei. 2011.
\newblock \href {https://doi.org/10.1504/IJKESDP.2011.039875} {Borderline
  over-sampling for imbalanced data classification}.
\newblock \emph{Int. J. Knowl. Eng. Soft Data Paradigm.}, 3(1):4–21.

\bibitem[{Nocedal(1980)}]{10.2307/2006193}
Jorge Nocedal. 1980.
\newblock \href {http://www.jstor.org/stable/2006193} {Updating quasi-newton
  matrices with limited storage}.
\newblock \emph{Mathematics of Computation}, 35(151):773--782.

\bibitem[{Reimers and Gurevych(2019)}]{reimers-gurevych-2019-sentence}
Nils Reimers and Iryna Gurevych. 2019.
\newblock \href {https://doi.org/10.18653/v1/D19-1410} {Sentence-{BERT}:
  Sentence embeddings using {S}iamese {BERT}-networks}.
\newblock In \emph{Proceedings of the 2019 Conference on Empirical Methods in
  Natural Language Processing and the 9th International Joint Conference on
  Natural Language Processing (EMNLP-IJCNLP)}, pages 3982--3992, Hong Kong,
  China. Association for Computational Linguistics.

\bibitem[{Sanh et~al.(2019)Sanh, Debut, Chaumond, and
  Wolf}]{Sanh2019DistilBERTAD}
Victor Sanh, Lysandre Debut, Julien Chaumond, and Thomas Wolf. 2019.
\newblock Distilbert, a distilled version of bert: smaller, faster, cheaper and
  lighter.
\newblock \emph{arXiv preprint arXiv:1910.01108}.

\bibitem[{Sia et~al.(2020)Sia, Dalmia, and Mielke}]{sia2020tired}
Suzanna Sia, Ayush Dalmia, and Sabrina~J. Mielke. 2020.
\newblock Tired of topic models? clusters of pretrained word embeddings make
  for fast and good topics too!
\newblock \emph{arXiv preprint arXiv:2004.14914}.

\bibitem[{Staykovski et~al.(2019)Staykovski, Barr{\'{o}}n{-}Cede{\~{n}}o,
  Martino, and Nakov}]{DBLP:conf/ecir/StaykovskiBMN19}
Todor Staykovski, Alberto Barr{\'{o}}n{-}Cede{\~{n}}o, Giovanni Da~San Martino,
  and Preslav Nakov. 2019.
\newblock \href {http://ceur-ws.org/Vol-2342/paper6.pdf} {Dense vs. sparse
  representations for news stream clustering}.
\newblock In \emph{Proceedings of Text2Story - 2nd Workshop on Narrative
  Extraction From Texts, co-located with the 41st European Conference on
  Information Retrieval, Text2Story@ECIR 2019, Cologne, Germany, April 14th,
  2019}, volume 2342 of \emph{{CEUR} Workshop Proceedings}, pages 47--52.
  CEUR-WS.org.

\bibitem[{Teh et~al.(2005)Teh, Jordan, Beal, and Blei}]{teh2005sharing}
Yee~W Teh, Michael~I Jordan, Matthew~J Beal, and David~M Blei. 2005.
\newblock Sharing clusters among related groups: Hierarchical dirichlet
  processes.
\newblock In \emph{Advances in neural information processing systems}, pages
  1385--1392.

\bibitem[{Weischedel et~al.(2013)}]{w13}
Ralph Weischedel et~al. 2013.
\newblock \emph{OntoNotes Release 5.0 LDC2013T19. Web Download}.
\newblock Linguistic Data Consortium, Philadelphia.

\bibitem[{Zaheer et~al.(2017)Zaheer, Ahmed, and Smola}]{pmlr-v70-zaheer17a}
Manzil Zaheer, Amr Ahmed, and Alexander~J. Smola. 2017.
\newblock \href {http://proceedings.mlr.press/v70/zaheer17a.html} {Latent
  {LSTM} allocation: Joint clustering and non-linear dynamic modeling of
  sequence data}.
\newblock In \emph{Proceedings of the 34th International Conference on Machine
  Learning}, volume~70 of \emph{Proceedings of Machine Learning Research},
  pages 3967--3976, International Convention Centre, Sydney, Australia. PMLR.

\bibitem[{Zaheer et~al.(2019)Zaheer, Ahmed, Wang, Silva, Najork, Wu, Sanan, and
  Chatterjee}]{10.1145/3289600.3291036}
Manzil Zaheer, Amr Ahmed, Yuan Wang, Daniel Silva, Marc Najork, Yuchen Wu,
  Shibani Sanan, and Surojit Chatterjee. 2019.
\newblock \href {https://doi.org/10.1145/3289600.3291036} {Uncovering hidden
  structure in sequence data via threading recurrent models}.
\newblock In \emph{Proceedings of the Twelfth ACM International Conference on
  Web Search and Data Mining}, WSDM ’19, page 186–194, New York, NY, USA.
  Association for Computing Machinery.

\bibitem[{Zhou et~al.(2015)Zhou, Xu, and He}]{zhou-etal-2015-unsupervised}
Deyu Zhou, Haiyang Xu, and Yulan He. 2015.
\newblock \href {https://doi.org/10.18653/v1/D15-1225} {An unsupervised
  {B}ayesian modelling approach for storyline detection on news articles}.
\newblock In \emph{Proceedings of the 2015 Conference on Empirical Methods in
  Natural Language Processing}, pages 1943--1948, Lisbon, Portugal. Association
  for Computational Linguistics.

\end{thebibliography}
\bibliographystyle{acl_natbib}

\appendix
\section{Appendix}
\label{sec:appendix}



The TDT corpus does not have a training and test split and we thus partition the corpus into two almost equal portions such that all documents in a single event are part of the same split. Our training set consists of 873 documents and our test set consists of 680 documents. The events in each partition of the TDT corpus is shown in Table \ref{table:tdt-events}

\begin{table}[h]
\small
\begin{tabular}{l}
\hline
\textbf{Events in Our Train Split}\\
\hline
Karrigan Harding, Shannon Faulker, Quayle lung clot,\\ 
Haiti ousts observers, NYC Subway bombing, \\ 
Carlos the Jackal, USAir 427 crash, Lost in Iraq, \\
Death of Kim Jong Il, Clinic Murders, Kobe Japan quake, \\
Serbs violate Bihac, OK-City bombing \\
\hline
\textbf{Events in Our Test Split}\\
\hline
Pentium chip flaw, Cuban riot in Panama, Justice-to-be Breyer, \\ 
Humble TX flooding, WTC Bombing trial, \\
Cessna on White House, Aldrich Ames, Comet into Jupiter, \\
Serbians down F-16, Carter in Bosnia, Halls copter, \\
DNA in OJ trial \\
\hline
\end{tabular}
\caption{Events in the training and test splits of the TDT Pilot corpus}
\label{table:tdt-events}
\end{table}

\appendix
\noindent
\textbf{Actual example of clusters incorrectly merged when documents are supplied out-of-temporal-order}. 
Cluster label \# 1024 in the Miranda test-set, contains articles on Qatar being selected as FIFA worldcup host and issues with immigrant labour there are discussed in negative sentiment. The ground truth is a large cluster with 1869 documents. 
An example document title in this cluster is ``Qatar World Cup sponsors targeted for improving workers' rights'' with timestamp 2015-05-25 15:27:00. 
Cluster \# 288 is a singleton about an upcoming Boston Celtics game and has a negative tone on their recent performance with an article titled ``Celtics kick away a winnable game'' with timestamp 2014-11-06 10:27:00. This is incorrectly merged with cluster \# 1024. There are many more clusters that are incorrectly merged with cluster \# 1024.

\end{document}